\documentclass[a4paper,fleqn]{cas-dc}
\usepackage{subcaption}
\usepackage[numbers]{natbib}
\usepackage[switch]{lineno}

\def\tsc#1{\csdef{#1}{\textsc{\lowercase{#1}}\xspace}}
\tsc{WGM}
\tsc{QE}
\tsc{EP}
\tsc{PMS}
\tsc{BEC}
\tsc{DE}

\begin{document}
\let\WriteBookmarks\relax
\def\floatpagepagefraction{1}
\def\textpagefraction{.001}

\shorttitle{Data-augmented Accident Anticipation}

\shortauthors{Yanchen Guan et~al.}

\title [mode = title]{World Model-Based End-to-End Scene Generation for Accident Anticipation in Autonomous Driving}                      


%
\author[1,2]{Yanchen Guan}\credit{Conceptualization of this study, Methodology, Experiment, Writing}
\author[1,3]{Haicheng Liao}\credit{Conceptualization of this study, Writing}
\author[1,2]{Chengyue Wang}\credit{Writing}
\author[1]{Xingcheng Liu}\credit{Experiment}
\author[1,2]{Jiaxun Zhang}\credit{Experiment}
\author[1,2,3]{Zhenning Li}[orcid=0000-0002-0877-6829]\cormark[1]\credit{Conceptualization of this study,Writing}\ead{zhenningli@um.edu.mo}


\cortext[cor1]{Corresponding author}

\affiliation[1]{organization={State Key Laboratory of Internet of Things for Smart City},
    addressline={University of Macau}, 
    city={Macau SAR},
    postcode={999078}, 
    country={China}}
\affiliation[2]{organization={Department of Civil Engineering},
    addressline={University of Macau}, 
    city={Macau SAR},
    postcode={999078}, 
    country={China}}
\affiliation[3]{organization={Department of Computer and Information Science},
    addressline={University of Macau}, 
    city={Macau SAR},
    postcode={999078}, 
    country={China}}

\begin{abstract}
Reliable anticipation of traffic accidents is essential for advancing autonomous driving systems. However, this objective is limited by two fundamental challenges: the scarcity of diverse, high-quality training data and the frequent absence of crucial object-level cues due to environmental disruptions or sensor deficiencies. To tackle these issues, we propose a comprehensive framework combining generative scene augmentation with adaptive temporal reasoning. Specifically, we develop a video generation pipeline that utilizes a world model guided by domain-informed prompts to create high-resolution, statistically consistent driving scenarios, particularly enriching the coverage of edge cases and complex interactions. In parallel, we construct a dynamic prediction model that encodes spatio-temporal relationships through strengthened graph convolutions and dilated temporal operators, effectively addressing data incompleteness and transient visual noise. Furthermore, we release a new benchmark dataset designed to better capture diverse real-world driving risks. Extensive experiments on public and newly released datasets confirm that our framework enhances both the accuracy and lead time of accident anticipation, offering a robust solution to current data and modeling limitations in safety-critical autonomous driving applications.

\end{abstract}




\maketitle

\section*{Introduction}

Autonomous driving technologies have rapidly advanced in recent years, with fully autonomous vehicles promising transformative improvements in road safety and transportation efficiency \cite{bagloee2016autonomous,kalra2016driving}. As a critical component of autonomous systems, accident anticipation aims to predict potential traffic accidents before they occur, allowing proactive interventions and enhancing driving safety \cite{fang2023vision,song2024dynamic}.

Since the concept of accident anticipation was proposed and its feasibility was demonstrated using dashcam videos and a dynamic spatial attention framework, researchers have approached this task from various perspectives \cite{chan2017anticipating,karim2022dynamic,malawade2022spatiotemporal}. By incorporating graph convolutional networks to model relationships among traffic participants and leveraging optical flow or depth features to enrich visual input, both the timeliness and accuracy of accident anticipation have improved substantially \cite{fatima2021global,yu2021scene,liu2023net,liao2024and,xiao2023review,thakur2024graph}.

Despite these promising advances, accident anticipation remains a profoundly challenging problem \cite{zhang2025latte,liao2024crash,liao2024real}. First, the limited visual cues captured by dashcams impose inherent constraints on feature extraction. Illumination changes, adverse weather, occlusions, and motion blur frequently obscure crucial spatiotemporal information-especially those related to vehicle dynamics-leading to information loss, where crucial spatial and temporal cues are either misinterpreted or entirely missed \cite{wang2024dc,cheong2017reflection}. 

Second, traffic accidents are inherently rare and highly diverse, posing challenges in collecting high-quality, large-scale video datasets \cite{sadeky2010real}. Existing datasets often exhibit significant discrepancies in accident types and driving contexts. For example, more than 60\% of accidents in the A3D dataset \cite{yao2019unsupervised} involve ego vehicles, whereas in the DAD dataset \cite{chan2017anticipating}, this figure drops to only 10\%. Such disparities hinder cross-domain generalization.

Furthermore, subtle environmental cues, such as nuanced vehicle behaviors or slight pedestrian movements, may serve as early indicators of imminent accidents. However, these signals are easily lost amid rapid scene transitions and temporal inconsistencies. Traditional approaches, though effective in some settings, often struggle to capture such fine-grained dynamics, particularly when relying on single-frame inputs \cite{wang2023gsc}.

\begin{figure}[htbp]
\centering  
\includegraphics[width=0.46\textwidth]{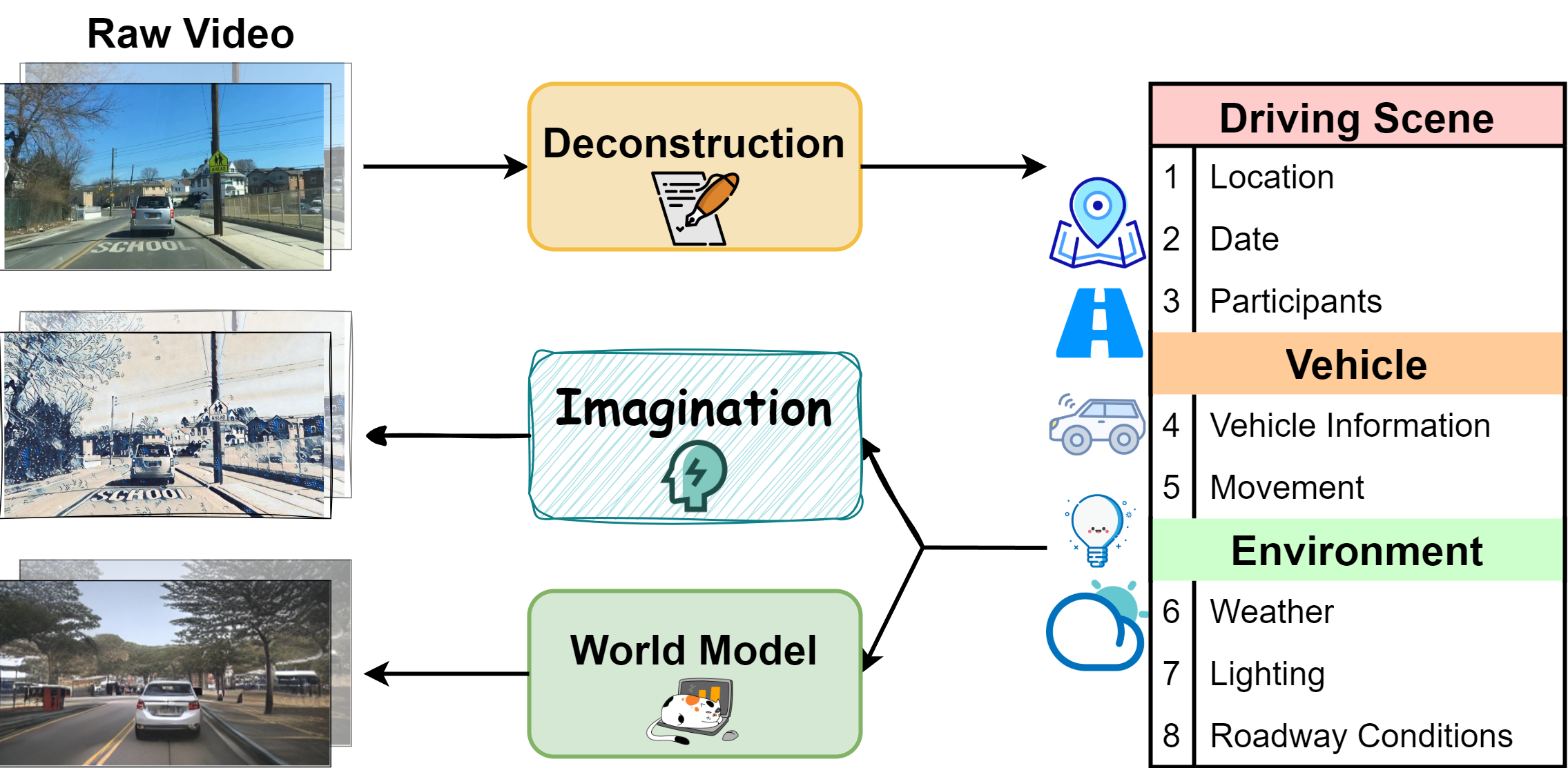}
\caption{Driving scene deconstruction and reconstruction diagram. For scene deconstruction, we use a vision-language model (VLM) to extract key domain knowledge through Zero-shot Visual Question Answering. In the process of scene reconstruction, humans interpret domain knowledge through common sense and imagine the scenes it constitutes . In our framework, a world model replaces the prior knowledge used in the human imagination process through the understanding of physical laws and common sense, and thus can complete the process of scene reconstruction.}
\label{Fig_first}
\end{figure}

To address these intertwined challenges, we propose a two-fold approach that simultaneously tackles data scarcity and modeling limitations from both the data generation and inference perspectives.

First, inspired by the structure of accident reports, we design a domain knowledge-guided framework to enrich the scale and diversity of training data using vision-language models (VLMs) and world models.
Recent progress in natural language processing \cite{devlin2019bert, zhang2024vision} has enabled the development of vision-language models that align visual and textual representations through large-scale pretraining \cite{radford2021learning}. Without requiring task-specific fine-tuning, such models can be directly applied to downstream visual tasks \cite{xiao2010sun}. In our work, we employ VLMs to extract diverse scene-level attributes from traffic videos. Unlike expert models designed for isolated tasks (e.g., weather or behavior recognition), a single VLM can simultaneously identify multiple semantic features by leveraging cross-modal correlations \cite{lohner2024enhancing}. Moreover, video-specific VLMs exploit temporal consistency across frames to mitigate errors from single-frame interpretation \cite{yang2022cross}.
In parallel, world models have emerged as powerful tools in autonomous driving, capable of modeling physical dynamics and generating realistic driving scenarios \cite{guan2024world, hu2023gaia, wang2024driving}. Unlike pure video generation models, world models incorporate physical laws to synthesize future scenes that mirror real-world conditions \cite{gao2024vista, liu2024sora, yang2024drivearena}.
Leveraging these capabilities, we extract accident-related domain features using VLMs to capture the original feature distribution, and employ a world model to generate synthetic driving scenes that preserve these distributions, as illustrated in Figure \ref{Fig_first}. This effectively augments the dataset with realistic, diverse, and high-fidelity samples-a critical step toward overcoming data scarcity \cite{ha2018recurrent, hafner2020mastering}. Nonetheless, care must be taken to ensure consistency between synthetic and real-world data.

Second, we develop a dynamic accident anticipation model based on enhanced graph convolutional networks (GCNs) with multi-layer dilated temporal convolutions. Unlike the fully connected graph neural networks commonly used in past research, this architecture captures complex spatial interactions and long-range temporal dependencies by adaptively modeling inter-agent relationships and expanding the receptive field over full video clips \cite{wang2023gsc,fang2023vision, thakur2024graph}. It significantly reduces the impact of transient information loss and improves the robustness of accident prediction in challenging, dynamic scenarios.

To validate our approach, we introduce the Anticipation of Traffic Accident (AoTA) dataset, which includes a large collection of real-world accident scenarios with rich annotations. We also benchmark several representative accident anticipation methods on AoTA. Experimental results show that our proposed model outperforms existing methods across multiple datasets and that the proposed data augmentation framework further enhances model performance and generalization.

Our main contributions are threefold:
\begin{itemize} 

\item We propose a driving scene generation framework for data augmentation in accident anticipation. This framework generates rich traffic scenes while preserving the original feature distributions, effectively addressing the data scarcity challenge. 

\item We introduce an accident anticipation model based on enhanced dynamic GCN and multi-layer dilated convolutions. By expanding the receptive field to process video clip features rather than single-frame inputs, our model reduces the occurrence of outliers and establishes new performance benchmarks across multiple datasets. 

\item We release the Anticipation of Traffic Accident (AoTA) dataset, which provides the largest set of accident cases and the most comprehensive annotations to date. Experimental results demonstrate state-of-the-art performance on this dataset. \end{itemize}

\section*{Methods}
\label{sec:Methodology}

The primary objective of this study is to predict the likelihood of an accident occurring based on dashcam video footage and to issue an early warning. Formally, let \( X = \{x^0, x^1, \dots, x^T\} \) represents a sequence of observed video frames, where \( T \) denotes the total number of frames. The model's task is to estimate the probability of an accident at each time step \( t \), represented as \( P = \{p^0, p^1, \dots, p^T\} \), where \( p^t \) is the predicted probability of an accident at time step \( t \). An accident is assumed to occur if the predicted probability first exceeds a predefined threshold \( p^{\tau} \) at a specific time step \( \bar{t} \), i.e., \( p^t \geq p^{\tau} \), where \( \bar{t} < T \). To evaluate the model's capability in predicting accidents in advance, we define the Time-to-Accident (TTA) as follows: \( \text{TTA} = \tau - \bar{t} \), where \( \tau \) represents the actual time of accident occurrence, with \( \tau = 0 \) for non-accident videos. A higher TTA value indicates the model's ability to anticipate accidents earlier, reflecting improved predictive performance. The objective of this work is to accurately estimate the probability sequence \( P \) of accident occurrences and to maximize the predicted lead time \( \bar{t} \) for videos containing traffic accidents.

\subsection*{Overall Framework}

\begin{figure}[t]
\centering  
\includegraphics[width=0.49\textwidth]{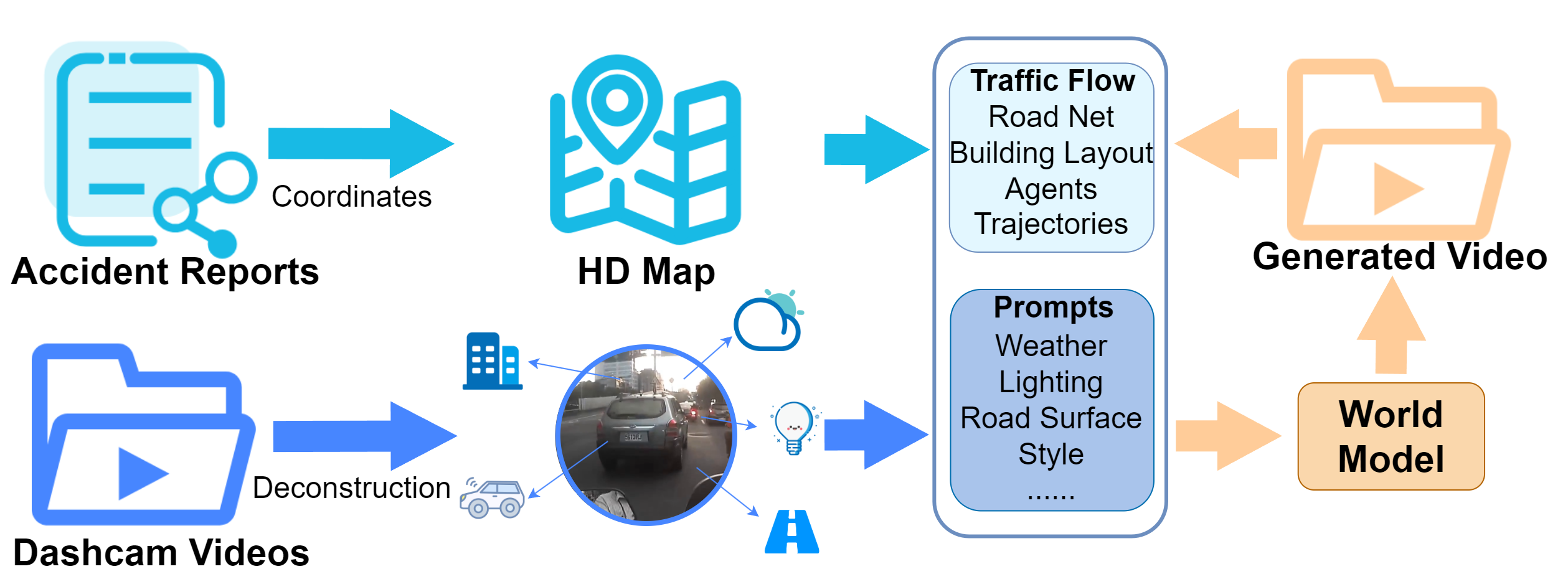}

\caption{Driving scenario generation framework. With the world model as the core, it uses scene factors and geographical conditions derived from real world data to generate realistic driving scenarios. All location coordinates are taken from real roadway traffic accident reports to ensure that the high-definition (HD) map used to generate the video is centered on the road.}
\label{Fig_Model}
\end{figure}

Our proposed framework consists of four key modules: 1) Driving Scene Generation, which utilize a world model to augment the training dataset by generate new driving scene from the origin dataset. 2) Visual Feature Extraction focuses on detecting traffic participants, estimating the video depth, extracting frame features and object features from the raw videos. 3) Dynamic Graph Conventional Network, which calculates the position of each target and uses a dynamic spatial graph neural network to dynamically model the interaction between targets. And 4) Temporal Relational Learning module is used to learn the temporal relationship between different video frames and use spatio-temporal features to predict the accident confidence probability of the video scene. In the following sections, we elaborate on the implementation of each module, while describing necessary details that cannot well presented above.

\begin{figure}[t]
\centering  
\includegraphics[width=0.46\textwidth]{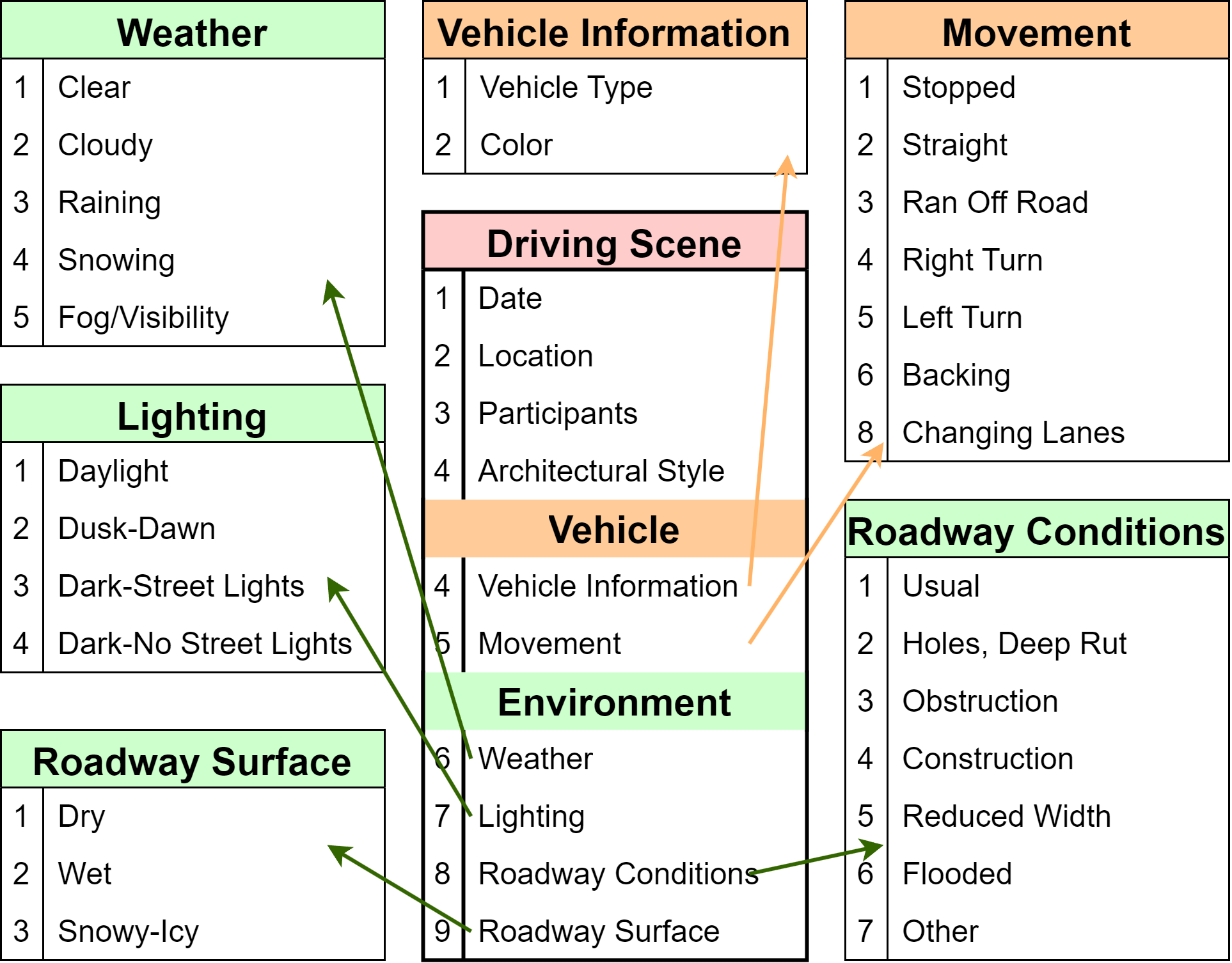}

\caption{Contributing factors in driving scene. These factors combined with driving common sense can be utilized to reconstruct the driving scenario in imagination.}
\label{Fig_scene_factors}
\end{figure}

\subsection*{Driving Scene Generation}
To address the lack of data diversity in existing accident anticipation datasets, we propose a framework for generating realistic style driving scenes as shown in Figure \ref{Fig_Model}. This framework aims to extract the domain knowledge distribution from existing driving scenes and reconstruct it into new driving scenes using a world model, ensuring that the newly generated driving scene data shares the same feature distribution as those in the dataset. Specifically, we first utilize Video-LLaVA \cite{lin2023video} to perform zero-shot visual question answering analysis on the videos in the dataset, extracting the scene feature distribution. Subsequently, we generate dashcam videos in a controllable manner by using random maps with the same feature distribution, and these generated videos are then mixed with the training data in the dataset for data augmentation. Since current research has not fully elucidated the visual feature clues in traffic accident scenes, in this paper, we only discuss the generation and usage of negative traffic scenes.

\subsubsection*{Driving Scene Deconstruction}

\begin{figure}[htbp]
\centering  
\includegraphics[width=0.46\textwidth]{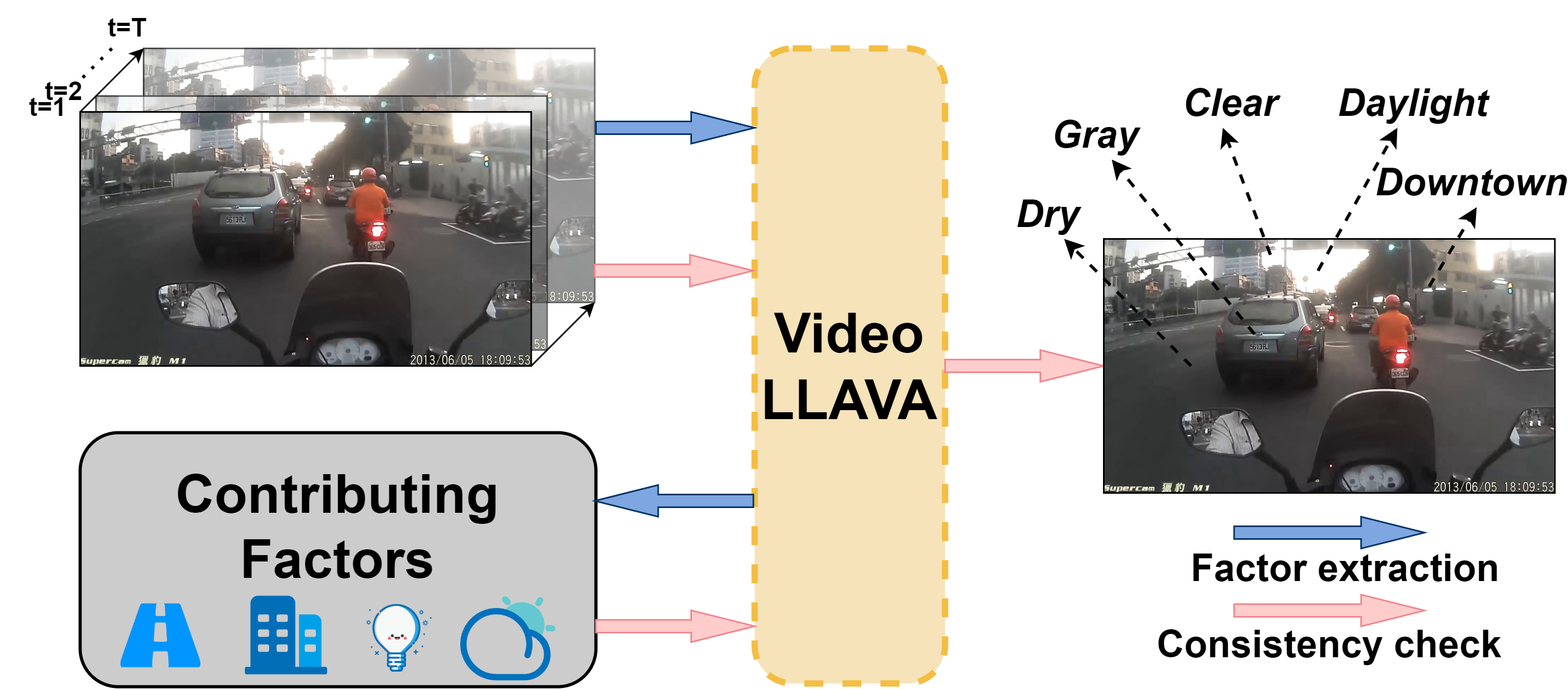}

\caption{Diagram of domain knowledge extraction and consistency check. Additional consistency checks on Vision-language model (VLM) output results through prompt engineering can effectively correct some incorrect answers. Samples that fail the consistency check will be revised through manual review.}
\label{Fig_Deconstruction}
\end{figure}

Figure \ref{Fig_scene_factors} illustrates key contributing factors related to visual features in driving scenes. To extract the visual contributing factors from the original video, we employ Video-LLaVA \cite{lin2023video}, which performs factor extraction and consistency checks through a fully automated process (see Figure \ref{Fig_Deconstruction}), thereby ensuring the accuracy of the obtained visual contributing factors. According to the evaluation on the CCD dataset, an accident anticipation dataset with annotations for lighting and weather conditions, the proposed pipeline achieves an average accuracy of 96.71\% in distinguishing daytime and nighttime conditions and 81.88\% in identifying weather conditions.

\begin{figure}[t]
\centering  
\includegraphics[width=0.36\textwidth]{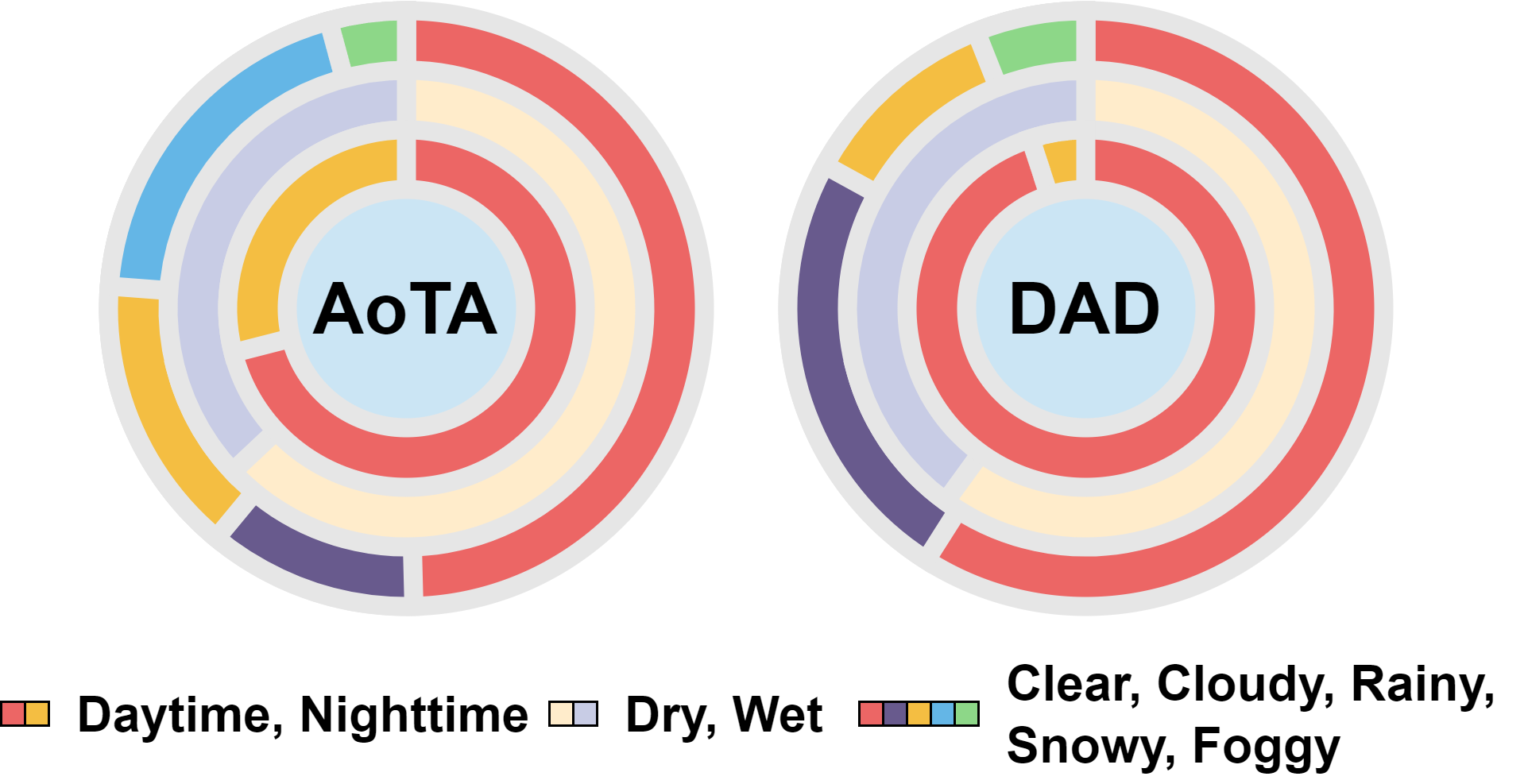}

\caption{Comparison of the environmental factor distributions of Anticipation of Traffic Accident (AoTA) dataset and Dashcam Accidents Dataset (DAD). The DAD dataset has a larger proportion of environments with good weather and high visibility, while the AoTA dataset is more balanced.}
\label{Fig_distribution}
\end{figure}

Considering the critical factors presented in Figure \ref{Fig_scene_factors}, it is possible to infer various environmental attributes, such as weather, lighting conditions, road status, and architectural style, by employing the scene deconstruction approach. Taking the AoTA dataset as an example, after decomposing all 1200 non-accident videos, we derived the proportion of different factors in this dataset, as illustrated in Figure \ref{Fig_distribution}. It can be seen that there are obvious differences in the distribution of environmental factors in the AoTA dataset and the DAD dataset, especially in the distribution of weather and lighting conditions, where AoTA shows higher diversity.
To maintain consistency in the distribution of data characteristics, the environmental factor distribution used as prompts in subsequent scene generation processes will retain the same proportion.

\subsubsection*{Driving Scene Reconstruction}

As illustrated in Figure \ref{Fig_Model}, our driving scene reconstruction framework is centered around a world model. The environment information obtained from driving scene deconstruction is used as prompts, combined with high-resolution maps and traffic flow data, to generate the complete driving environment. The driving agent takes the generated driving environment as input and outputs actions and trajectory planning, which are used to update the driving scene information, forming a closed-loop driving scene generation system.

\paragraph{\textbf{Locating and route planning}}

Determining the location for driving scenes is the first step in generating new driving scenarios. To achieve efficient large-scale automated generation, we randomly select coordinates from publicly available traffic accident reports released by the state of Maryland, USA, as virtual coordinates for new scenes. By utilizing coordinates from traffic accident reports, we ensure that the designated driving scene locations are on or near roads, avoiding the issue of generating unrealistic driving scenes far from roadways.

Our scene generation framework allows for custom scene generation at arbitrary locations in any city using OpenStreetMap \cite{haklay2008openstreetmap}, facilitating the construction of diverse road networks and driving environments. We import the selected coordinates into OpenStreetMap and download all road and building information within a 500-meter square area, enabling our generated scenes to correspond to real-world environments and ensuring that the road and building layouts maintain a high level of realism.

Subsequently, SUMO \cite{krajzewicz2012recent} is used to randomly generate traffic participants. By randomly controlling the number of traffic participants within a certain range and assigning each participant a random start and destination point on the road network, the traffic flow in the new scene can be reasonably generated based on the road network information.

After obtaining global traffic flow, we focus on the precise control of the ego vehicle. UniAD \cite{hu2023planning}, a representative end-to-end driving agent, is employed as the driving agent for vehicles within the generated scenes. UniAD utilizes vision-only inputs to predict the motion of both ego and surrounding vehicles, subsequently outputting trajectory planning.

\paragraph{\textbf{Scene generation}}

After the data preparation stage, all necessary information for constructing new driving scenes is obtained, including environment prompts (e.g., weather, lighting, road conditions, environmental styles) randomly generated based on the feature distribution of the dataset, HD maps, traffic flow, as well as ego-vehicle motions and trajectory planning.

An advanced world model based on a stable diffusion pipeline trained on nuScenes \cite{caesar2020nuscenes} and initialized with the pre-trained Stable Diffusion v1.5 \cite{rombach2022high}, known as World Dreamer \cite{yang2024drivearena, wang2024worlddreamer}, is used for the scene generation. World Dreamer utilizes an efficient condition encoding module to accept traffic flow, vehicle layouts, prompts, and other inputs, and generates high-quality images. To maintain temporal consistency, World Dreamer uses previously generated images and the corresponding ego-pose as references, guiding the model to generate the current frame while enhancing consistency between consecutive frames.

\begin{figure}[htbp]
\centering

\includegraphics[width=0.48\textwidth]{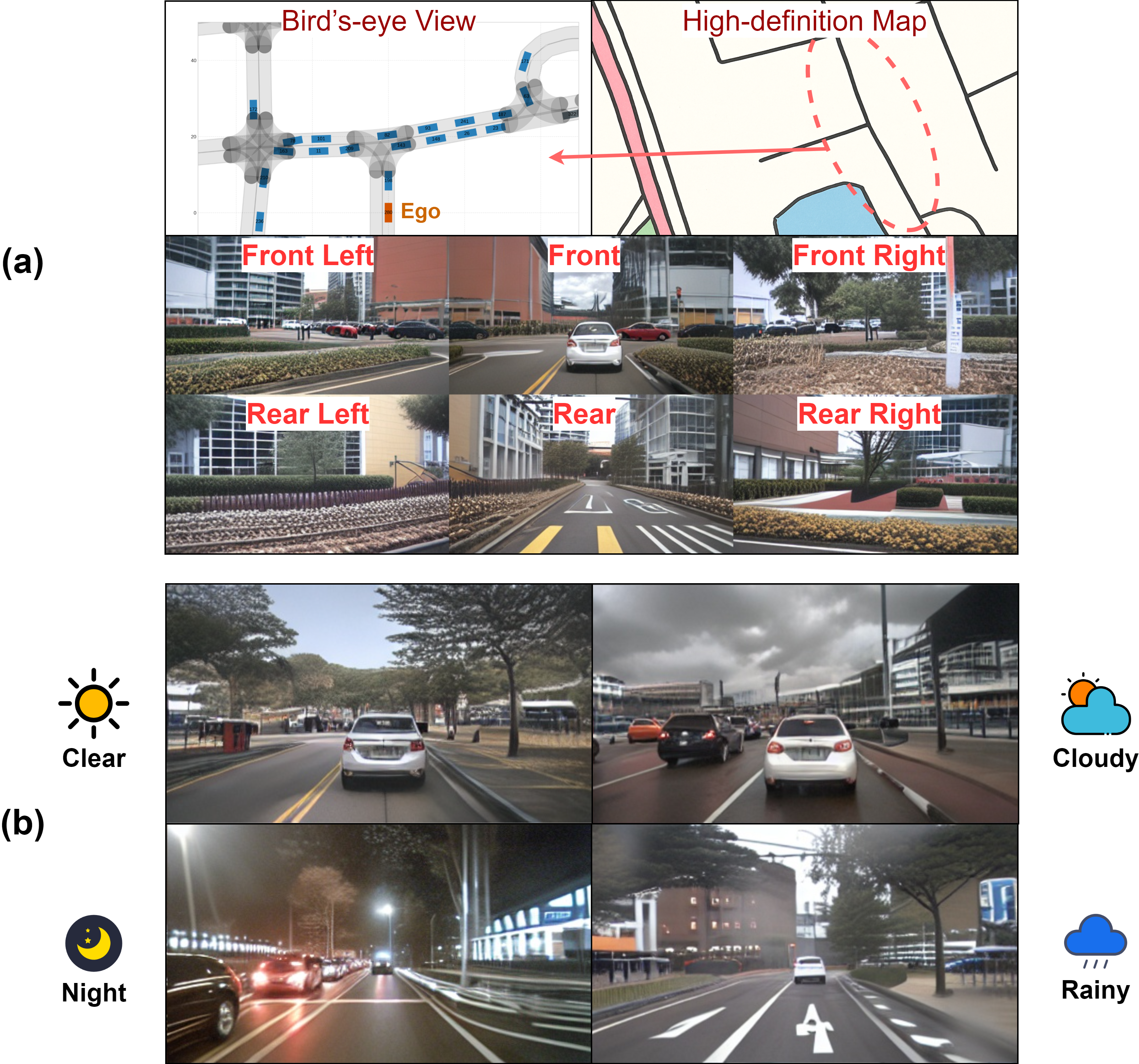}

\caption{Examples of generated videos. \textbf{a} Generated video with high-definition (HD) map and bird’s-eye view (BEV). \textbf{b} Generated video with different weather.}
\label{BEV_weather}
\end{figure}

The step size of World Dreamer is set by taking the FPS of the original video as a reference, the width and height of the output image are set to be consistent with the original video, and we summarize the images from the front camera into the generated dashcam videos. Figure \ref{BEV_weather}a shows a high-definition map of a traffic scene, vehicle occupancy from the BEV’s perspective, and the final driving video. And Figure \ref{BEV_weather}b shows the generated videos with diverse weather conditions. As can be seen from the examples, although the generated videos have a realistic style and are generally logical, they still have some shortcomings. Compared with real driving videos, the clarity is still not enough, the consistency of lane lines and the environment cannot be guaranteed over a long time span, and there is a certain degree of deformation of vehicles and buildings. Despite these minor shortcomings, the generated videos are able to describe a complete traffic scenario in terms of environment and vehicle operation, and can meet the needs of serving as accident anticipation training data.

To evaluate the quality of the generated videos, we utilize Fréchet Video Distance (FVD) \cite{unterthiner2019fvd} (similar to FID \cite{yu2021frechet}) as the primary metric, using the DAD dataset as the reference. The FVD metric assesses both visual quality and spatiotemporal consistency. The experimental results indicate that the generated video set achieves an FVD score of 36.38 on the DAD dataset, suggesting that although the visual quality of the generated content still exhibits certain limitations, the distributional discrepancy between the generated and real videos in the feature space remains relatively minor when considering spatiotemporal consistency. This observation aligns with the performance of our GNN-based accident prediction model, which leverages object distance and velocity weights for inference.

 \begin{figure*}[ht]
\centering  
\includegraphics[width=0.75\textwidth]{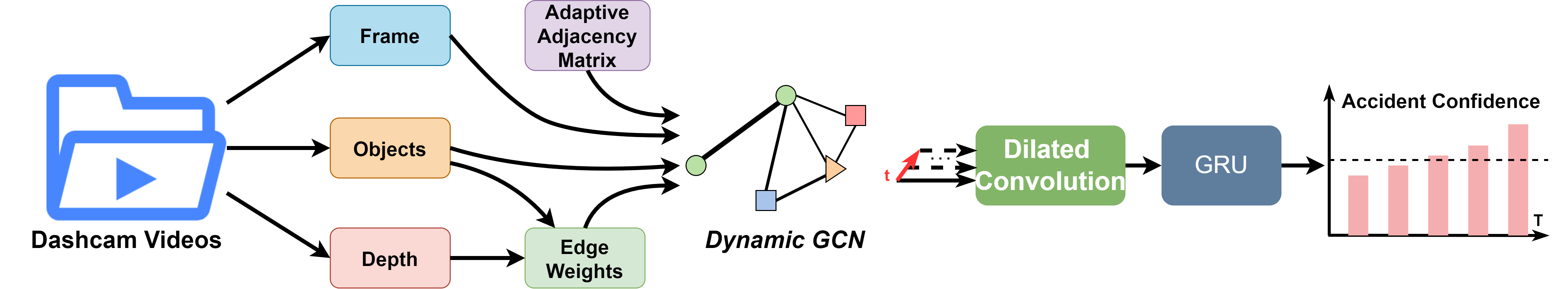}

\caption{Our proposed traffic accident anticipation framework. This framework includes three parts: extracting visual features and depth information from dashcam videos, using dynamic graph neural network (GCN) to model object relationships, and long-term temporal learning.}
\label{GCN}
\end{figure*}

\subsection*{Visual Feature Extraction}

Figure \ref{GCN} shows the overall framework of our proposed traffic accident anticipation model.
For each frame in dashcam videos, we first employ an object detection network to obtain all candidate objects \cite{ouyang2022nms}. To eliminate redundant bounding boxes, we utilize the Intersection over Union (IoU) metric among different boxes and retain the top 19 objects with the highest confidence scores. Subsequently, we extract visual features from video frames and all candidate object boxes using the VGG16 network \cite{simonyan2014very} trained on ImageNet-1K. Furthermore, we estimate the depth value in the video frame using a depth estimation network (ZOE Depth trained on  KITTI \cite{bhat2023zoedepth, menze2015object}).

\begin{equation}
\begin{aligned}
&X \in \mathbb{R}^{B \times T \times N \times F}
\\
&Det \in \mathbb{R}^{B \times T \times (N-1) \times F}
\\
&Depth \in \mathbb{R}^{B \times W \times H}
\\
\end{aligned}
\end{equation}  
where, $X$, $Det$ and $Depth$ represents feature data, object detection data and depth data respectively. And $B$, $T$, $N$ and $F$ represents batch size, number of frames, number of total features of both frame and objects and feature dimension. Besides, for the $Depth$, $W$ and $H$ denotes for the width and height of the video.

\subsection*{Dynamic Graph Convolutional Network}

The spatial relationships among traffic participants at each moment provide sufficient information for accident anticipation. In this module, we employ an improved Dynamic GCN to model the spatial interactions between agents in traffic scenarios. By representing the observations at each time step using a graph structure, the model effectively minimizes the influence of background noise in the video and highlights the important interaction relationships among the traffic participants.

First, we treat the detected objects as graph nodes, forming a complete graph structure at each time step. Since it is not immediately clear which objects will be involved in an accident, the adaptive adjacency matrix between all graph nodes is initially set as fully connected. The adjacency matrix is defined as:
\begin{equation}
A = \phi_{\textit{softmax}}(V_1 \cdot V_2)
\end{equation}
where, $A \in \mathbb{R}^{B \times T \times N \times N}$, $V_1\in \mathbb{R}^{B \times N \times N}$ and $V_2\in \mathbb{R}^{B \times N \times N}$ are learnable matrices.

Generally, agents that are in close proximity to others have a higher likelihood of being involved in future accidents, and those with higher relative velocities are also more prone to collisions. Conversely, agents that are moving away from each other have a lower probability of being involved in an accident. Therefore, both distance and relative velocity between nodes should be considered in edge weight calculations.

\begin{figure}[t]
\centering  
\includegraphics[width=0.46\textwidth]{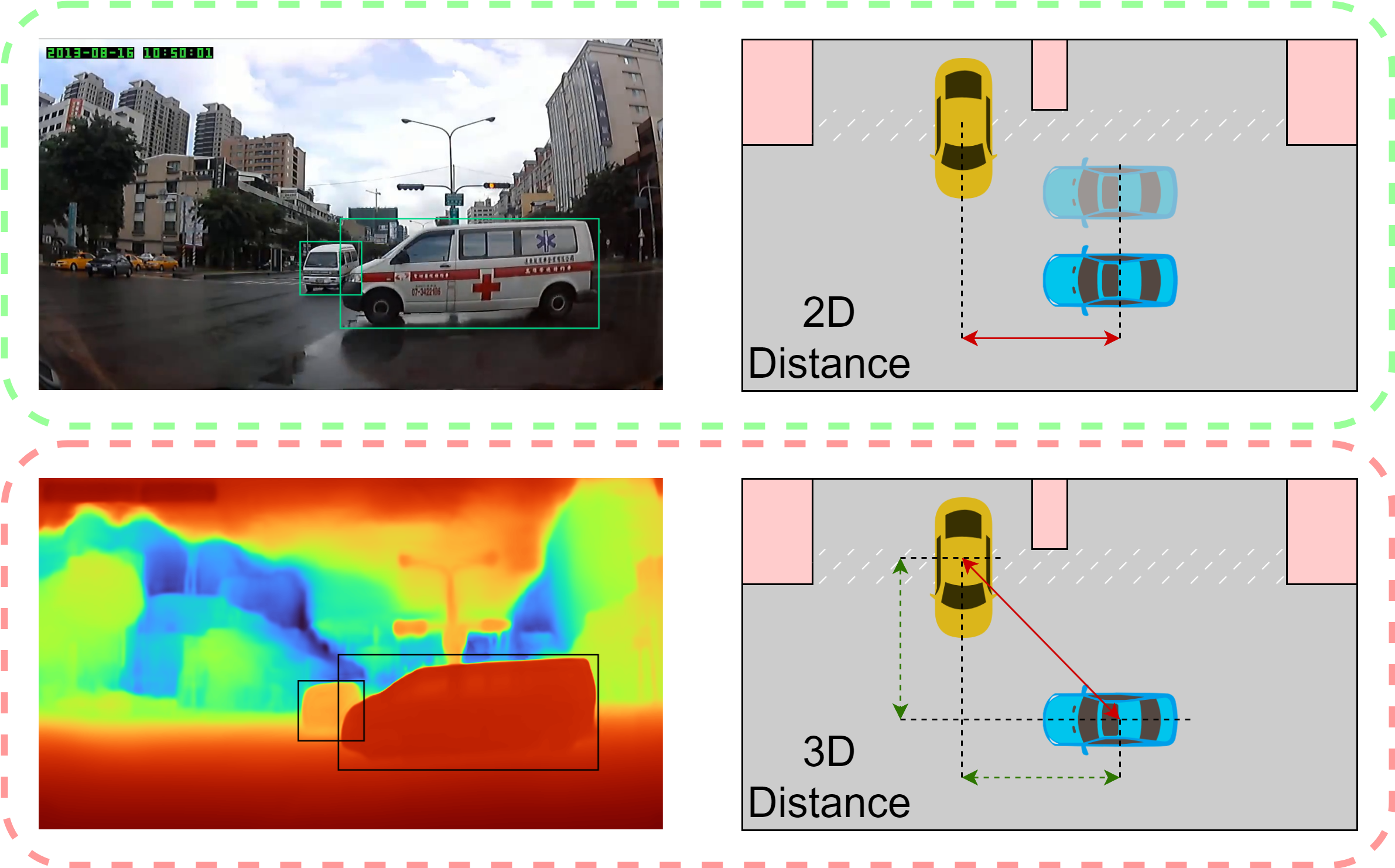}

\caption{Comparison of 2D distance and 3D distance in distance estimation between objects. When use 2D distance, one distance corresponds to multiple positional relationships. After adding depth information to calculate the 3D distance, the positional relationship becomes unique.}
\label{distance}
\end{figure}

Traditional methods often use the pixel distance between the bounding box centers of two agents as a reference for distance \cite{wang2023gsc, thakur2024graph}. However, since dashcams are monocular, the 2D pixel distance cannot accurately reflect the vertical distance between the agents, leading to errors in edge weights. We incorporate the depth difference between agents when calculating their distance. As shown in Figure \ref{distance}, the computed Euclidean distance more accurately represents the actual distance between agents.

For any target $i$, its pixel position $C_i$ is given by:
\begin{equation}
C_{ix} = \frac{x_{\text{min}} + x_{\text{max}}}{2}, \quad C_{iy} = \frac{y_{\text{min}} + y_{\text{max}}}{2}
\end{equation}
where $x$ and $y$ represent the corresponding bounding box coordinates in the detection dataset. Thus, the pixel distance $\text{Dist}_{ij}$ and depth difference $\text{Depth}_{ij}$ between two agents $i$ and $j$ are defined as:
\begin{equation}
\begin{aligned}
&\text{Dist}_{ij} = \frac{1}{D_d} \cdot \lvert C_i - C_j \rvert \\
&\text{Depth}_{ij} = \lvert D_i - D_j \rvert \\
\end{aligned}
\end{equation}
where, $D_d$ is the diagonal pixel distance, which is used as a normalization factor. For a 720*1280 video, the value of $D_d$ is 1450. Distance $D_{ij}$ between two agents and their relative velocity $\text{Vel}_{ij}(t)$ at time $t$ are given by:
\begin{equation}
\begin{aligned}
&\text{D}_{ij} = \sqrt{\text{Dist}_{ij}^2 + \text{Depth}_{ij}^2} \\
&\text{Vel}_{ij}(t) = \text{D}_{ij}(t) - \text{D}_{ij}(t - 1) \\
\end{aligned}
\end{equation}

Since the optimal weighting ratio between distance and relative velocity for graph construction is not known, we introduce an adaptive parameter $a$, which is learned by the model during training to determine the edge weight:
\begin{equation}
\text{Weight}_{ij} = \frac{a}{a + 1} \cdot e^{-\text{D}_{ij}} + \frac{1}{a + 1} \cdot \text{Vel}_{ij}
\end{equation}
The initial value of $a$ is set to 1.

As shown in Figure \ref{GCN}, we construct a weighted graph convolutional layer using node features, edge weights, and the adjacency matrix:
\begin{equation}
H^{(l+1)} = \sigma\left(\tilde{A} \cdot H^{(l)} \cdot W^{(l)}\right)
\end{equation}
where $H^{(l)}$ represents the input features at layer $l$, $W^{(l)}$ is the weight matrix of that layer, $\tilde{A}$ is the normalized adjacency matrix, and $\sigma$ denotes the ReLU activation function. Finally, we use LSTM to capture temporal dependencies across consecutive time steps, yielding the hidden state $h_t$:
\begin{equation}
h_t, c_t = \phi_{\textit{LSTM}}(x_t, (h_{t-1}, c_{t-1}))
\end{equation}
where, $x_t$ is the input of LSTM layer at time $t$. 

At this stage, we obtain graph features representing the spatial relationships among agents, which are subsequently fused with global visual features for further temporal relationship learning.

\subsection*{Temporal Relational Learning}

To enable the model to learn temporal relationships across different time steps, existing works have widely adopted methods such as LSTM \cite{hochreiter1997long}, GRU \cite{chung2014empirical}, and TCN \cite{lea2017temporal}. However, these methods process information from only a single time step at a time. Due to inconsistencies in identifying specific objects across all time steps and changes in the graph structure caused by objects entering or leaving the frame, relying solely on single time-step information often leads to disruptions. Therefore, we propose using multi-layer dilated convolutions \cite{yu2015multi} to expand the receptive field of the model in each step, which can effectively solve the problems of information loss in individual frames and occasional outliers generated by the model, thereby addressing the temporal continuity issue.

We first transpose the input tensor to $X' \in \mathbb{R}^{B \times F \times T}$, where the convolution output at layer $i$ is given by:
\begin{equation}
Y_i(t) = \sum_{k=0}^{K-1} W_i[k] \cdot X'(t - r_i \cdot k)
\end{equation}
where $Y_i(t)$ represents the convolution output at time step $t$ for layer $i$, $W_i[k]$ denotes the $k$-th weight of the convolution kernel at layer $i$, $r_i$ is the dilation rate for layer $i$, and $k$ is the kernel size. After three layers of dilation, the receptive field for each time step of the model expands to 8 steps. The final output tensor is:
\begin{equation}
X'' = \sigma\left(Y_L \left( \sigma\left( Y_{L-1} \left( \dots \sigma\left( Y_1(X') \right) \dots \right) \right) \right) \right)
\end{equation}

After passing through all convolution layers, a skip connection is applied by adding the input to the output, followed by layer normalization. The output of the multi-layer dilated convolutions is:
\begin{equation}
Z = \phi_{\textit{LayerNorm}}(X'' + X)
\end{equation}

Finally, we use a GRU for long-term temporal relationship learning, and the prediction results are obtained through linear classification:
\begin{equation}
\text{Out} = \phi_{\textit{GRU}}(Z), \quad \text{Out} \in \mathbb{R}^{B \times F \times 2}
\end{equation}

The loss function of our model is a time-weighted cross-entropy function between the predicted results and the video labels:
\begin{equation}
\begin{aligned}
&L_{\text{CE}} = -\sum_{i} y_i \log(\text{Out}_i)\\
&L = \exp\left(-\frac{(\text{toa} - t - 1)}{\text{fps}}\right) \cdot L_{\text{CE}}+L_{\text{CE}}\\
\end{aligned}
\end{equation}
where, for the $t$-th frame, $y_i$ is the ground truth label, $\text{Out}_i$ is the predicted result, and $\text{toa}$ is the actual time of the accident.

\subsection*{Anticipation of Traffic Accident Dataset}

In this paper, we introduce the Anticipation of Traffic Accident (AoTA) dataset, a challenging benchmark for the accident anticipation task. The AoTA dataset is curated from the Detection of Traffic Anomaly dataset \cite{yao2022dota} and BDD100K \cite{yu2020bdd100k}, designed to provide a diverse and extensive resource for advancing research in this domain.

\begin{figure}[t]
\centering  
\includegraphics[width=0.48\textwidth]{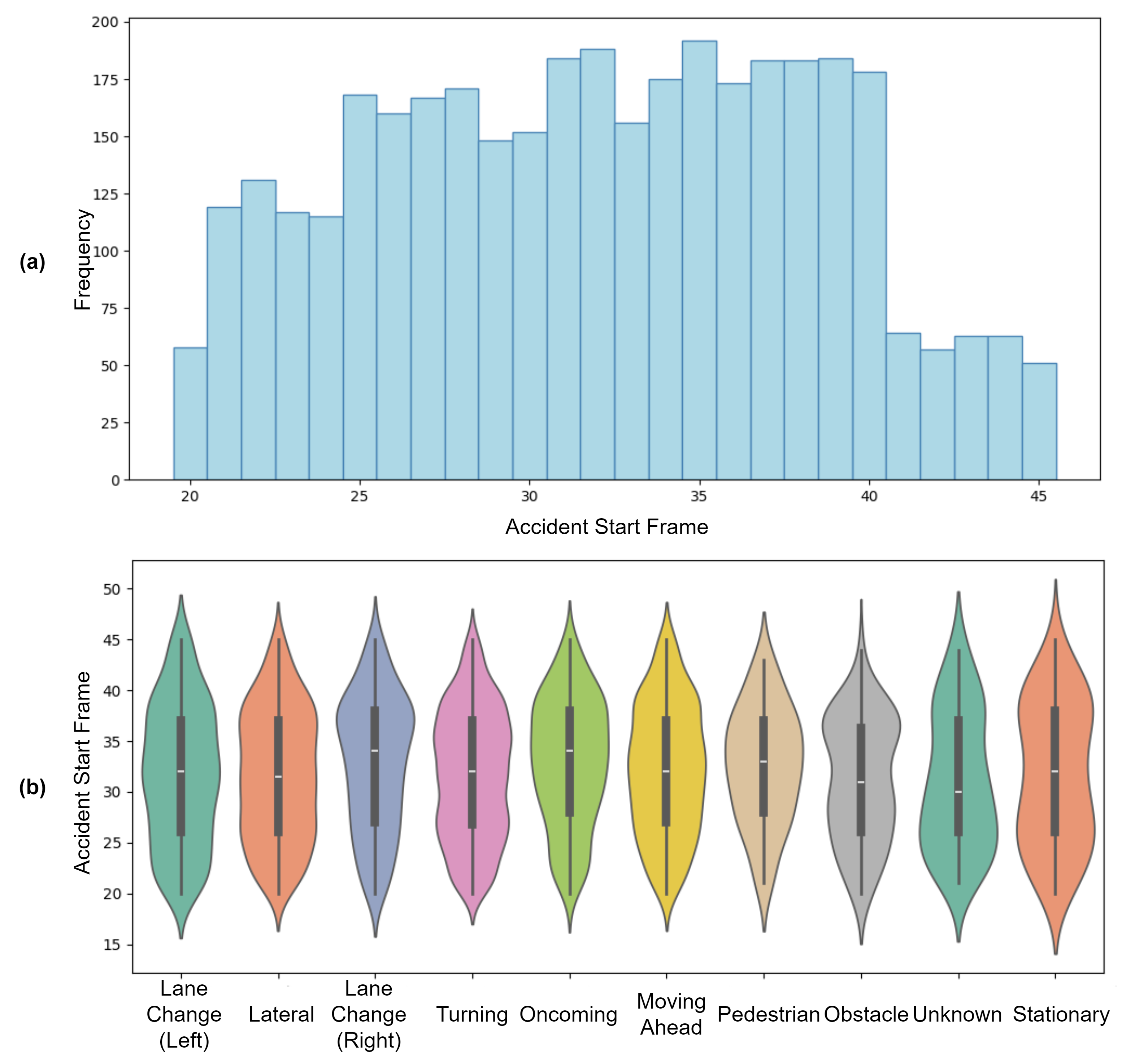}
\caption{Distribution of traffic accident start frames in the Anticipation of Traffic Accident (AoTA) dataset. \textbf{a} Histogram showing the frequency of all accident start frames across positive samples in the AoTA dataset. \textbf{b} Violin plots showing the distribution of accident start frames across different accident types. For each category, the violin shape represents the kernel density estimate (KDE) of start frames, while the embedded boxplot indicates the median (white dot), interquartile range (black bar), and the whiskers extending to 1.5xIQR (interquartile range). In the AoTA dataset, the time of traffic accidents is completely random and there is no obvious difference in the distribution of accident start times between different accident types, thus simulating naturally occurring random traffic accidents.}
\label{Aota_start}
\end{figure}

The AoTA dataset is constructed by selecting 4,000 accident clips from the Detection of Traffic Anomaly dataset, we edit and re-annotate those videos to mark the precise moment of accident occurrence. Following this process, a total of 3,600 accident videos are retained as positive samples after further filtering based on the time of accident occurrence. In these positive videos, the accident typically occurs between frames 20 and 45, Figure \ref{Aota_start} shows the frequency of different traffic accident start times in the dataset and the distribution of accident start times corresponding to different accident types. Negative samples consist of 1,200 normal driving dashcam clips from the BDD100K dataset, covering common traffic scenarios.

The AoTA dataset is divided into a training set containing 4,000 videos and a test set consisting of 800 videos. Table \ref{datasets} provides a detailed comparison between AoTA and existing datasets. Compared to other datasets, AoTA not only includes more samples but also features a wider range of driving environments and richer annotations.

\paragraph{\textbf{AoTA+}}

Based on AoTA, we apply the proposed driving scene generation framework to analyze all negative videos, as illustrated in Figure \ref{Fig_distribution}. Under the condition of maintaining similar feature distributions, the driving scene generation framework generates 300 entirely new negative videos, which are then added to the AoTA training set. This augmentation increases the number of negative videos from 1,000 to 1,300, resulting in an enhanced dataset, referred to as AoTA+.

To evaluate the effect of data augmentation using the generated videos, we deploy multiple baseline models on both AoTA and AoTA+ while keeping the test set identical. The results demonstrate the impact of the generated negative samples on model performance, as shown in Table \ref{table:balance}.

\subsection*{Experimental Setup}
\label{sec:Experiments}

All models are implemented using PyTorch and executed on an RTX 4090 GPU for both training and testing. The dimensionality of the extracted features for both detection boxes and video frames are set to 512. During the training phase of our model, we employ the ReduceLROnPlateau scheduler for learning rate adjustment, with a factor set to 0.5 and a patience of 3 epochs, training for a total of 15 epochs, with the batch size set to 10. The Adam optimizer is utilized with an initial learning rate of $1 \times 10^{-4}$, L1 and L2 regularization are applied to prevent over fitting, with weight decay set to $1 \times 10^{-3}$ and $1 \times 10^{-4}$. In the visualization stage, we use Gaussian smoothing to process the output results, and the sigma value is set to 2.

\subsubsection*{Datasets}

We evaluate the performance of our model on three real-world datasets: the Dashcam Accidents Dataset (DAD) \cite{chan2017anticipating}, the AnAn Accident Detection (A3D) \cite{yao2019unsupervised} dataset, the Anticipation of Traffic Accident (AoTA) dataset, and a dataset augmented with generated data: AoTA+. 

\paragraph{\textbf{DAD}}
The Dashcam Accidents Dataset (DAD) \cite{chan2017anticipating}, introduced by Chan et al., consists of 1,750 dashcam videos collected from six cities in Taiwan. This dataset includes 620 accident videos and 1,130 non-accident videos. The videos are split into 1,284 training samples (455 positive and 829 negative) and 466 testing samples (165 positive and 301 negative). Each video spans 5 seconds and contains 100 frames, with accidents in positive samples occurring at the 90th frame.

\paragraph{\textbf{A3D}}
The An An Accident Detection (A3D) dataset \cite{yao2019unsupervised} comprises 1,500 dashcam video clips from East Asian urban environments, covering diverse weather conditions and times of day. The features utilized in this work are provided by Bao et al. \cite{BaoMM2020}. Captured at 20 FPS, each clip includes 100 frames and is divided into training and testing sets with an 80\%-20\% split. Designed for traffic anomaly detection tasks, the A3D dataset contains only positive samples. Negative samples are extracted from video segments without accidents, enabling the dataset to support traffic accident anticipation tasks.

\subsubsection*{Evaluation Metrics}
In the traffic accident anticipation task, the most important indicators are Average Precision (AP) and Mean Time-To-Accident (mTTA).

\paragraph{\textbf{Average Precision (AP)}} is a metric used to evaluate the accuracy of predicting results. AP measures the model's ability to correctly identify the occurrence of traffic accidents, particularly in cases with an imbalance between positive and negative samples. This is done by assessing both precision and recall across various threshold settings.

In binary classification tasks, precision (P) is the ratio of true positives (TP) to the sum of true positives (TP) and false positives (FP), while recall (R) is the ratio of true positives (TP) to the sum of true positives (TP) and false negatives (FN). The precision-recall curve is plotted using these values, and AP is defined as the area under this curve.

For a given threshold $p$, if the confidence score $a_{t}^{p}$ at time $t$ exceeds this threshold, a traffic event is predicted. The resulting classifications are categorized as TP, FP, FN, and TN. The AP is then calculated by integrating the precision-recall curve, which can be approximated by discrete summation:

\begin{equation}
AP = \int P(R) \, dR = \sum_{k=0}^{m} P(k) \Delta R(k)
\end{equation}

\paragraph{\textbf{Mean Time-To-Accident (mTTA)}} quantifies a model's ability to predict the occurrence of an accident among positive samples in advance. This metric evaluates the earliness of accident anticipation based on positive predictions from dashcam videos. 

If an accident is predicted to happen at the $x$-th frame and actually occurs at the $y$-th frame, the time-to-accident (TTA) is defined as $y$-$x$.  By analyzing different threshold values $p$, sequences of TTA results and corresponding recall rates can be derived. The average of these TTA values is termed mTTA. Nonetheless, it is crucial to recognize that a high TTA might be misleading if the model overfits the dashcam data, resulting in indiscriminate positive predictions. Hence, a high TTA value is ineffective without considering the AP. This study reports the TTA values when the highest AP is attained, ensuring a comprehensive evaluation of the model's predictive performance.

\begin{figure*}[ht]
\centering  
\includegraphics[width=0.65\textwidth]{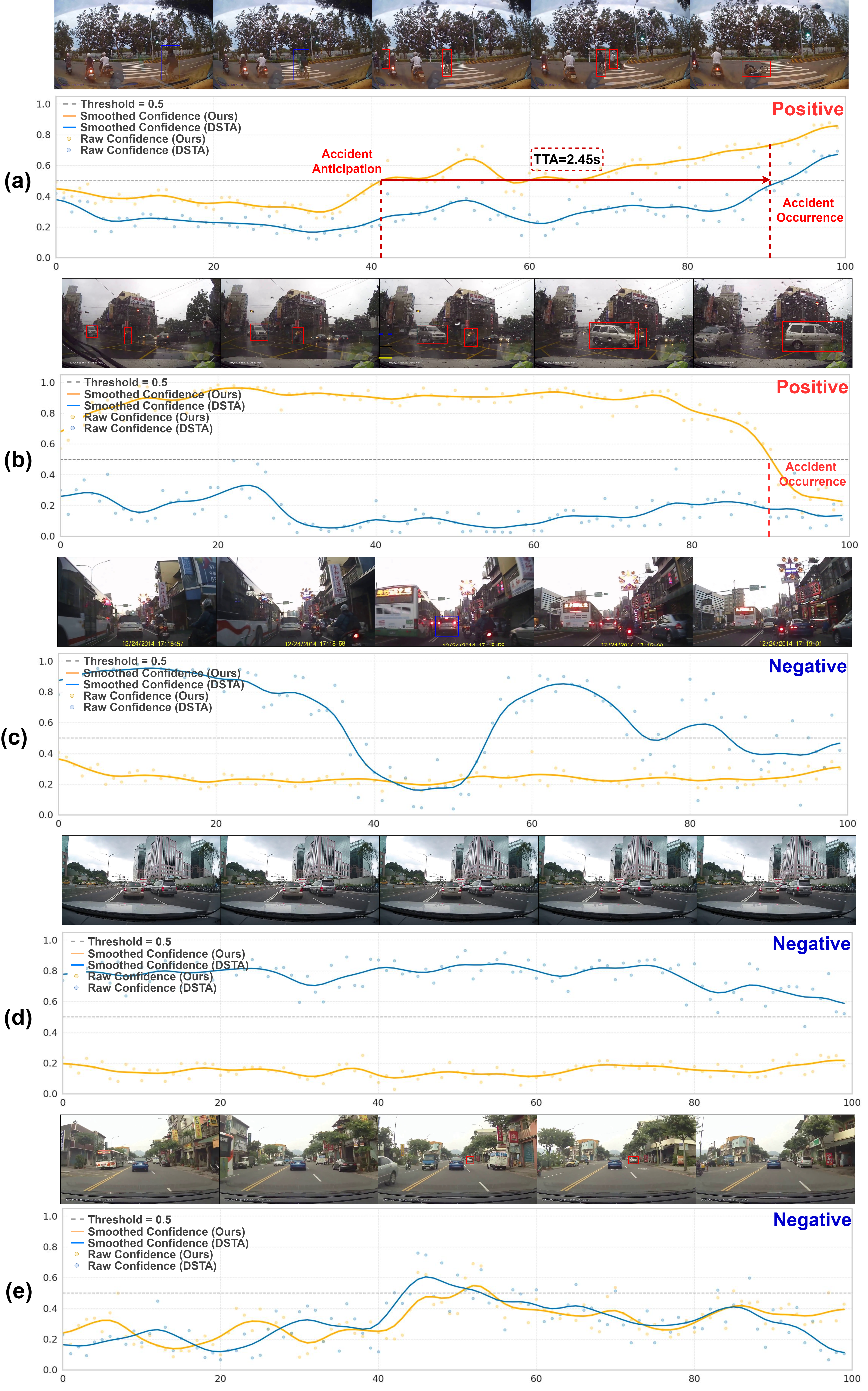}
\caption{Visualization of our model performance and comparison with another representative model DSTA.  \textbf{a, b} Positive scenarios where our anticipation results differ from those of DSTA. \textbf{c, d} Negative scenarios where our anticipation results differ from those of DSTA. \textbf{e} A scenario of anticipation failure.}
\label{Fig_visual}
\end{figure*}

\section*{Results}

Table \ref{table:balance} presents the top performance of our model on the DAD, A3D, AoTA and AoTA+ datasets. Compared to previous state-of-the-art methods, our model achieved an improvement of 7.0\% in AP and 9.1\% in mTTA on the DAD dataset. Moreover, on the A3D dataset, despite the current AP approaching saturation, our model still attained superior performance. Additionally, when focusing solely on the AP metric, our model consistently maintains a leading performance.

In our newly proposed AoTA and AoTA+ datasets, we reproduced several representative frameworks and compared their performance with our model. On the AoTA dataset, our model achieved the best performance, surpassing the second-best model by 3.9\% in AP and 33.2\% in mTTA. Similarly, on the AoTA+ dataset, we obtained comparable advantages. By comparing the performance of each framework on AoTA and AoTA+, it can be observed that the introduction of synthetic video data in the training set led to slight improvements for all models, with an average AP increase of 2.5\%, and a maximum improvement of 4.4\%, while the mTTA remained close to the original performance. This demonstrates that augmenting the original dataset with dashcam videos generated by our proposed driving scene generation framework effectively enhances the model's prediction accuracy for accident detection.

Furthermore, we conducted experiments by incrementally incorporating augmented data into the existing datasets to investigate the impact of integrating different proportions of generated data into the training set on model performance. As shown in Table \ref{table:dad+}, we augmented the DAD dataset's training data by adding between 10\% and 40\% of negative samples, as well as directly replacing 40\% of the negative sample data. The experimental results show that as the generated traffic scene data is added to the training set, the performance of the model gradually improves with the increase of input data. However, when we directly use the generated data to replace part of the source data, the performance of the model declines. Combined with the experimental results in AoTA and AoTA+, it can be seen that the generated traffic scene data is helpful for accident prediction tasks, but there is still a certain gap in the quality of generated traffic scene compared with the original videos.

\subsection*{Ablation Studies}

Table \ref{Ablation} shows the results of our ablation study on the DAD dataset, evaluating the impact of key modules, including GRU, multi-layer dilated convolution, dynamic GCN, and adaptive adjacency matrix, on the model's performance. 
\begin{itemize}

  \item \textbf{Model A} Removing the GRU led to decreases in both AP and mTTA, highlighting the importance of learning temporal relationships across frames. 
  \item \textbf{Model B} When the multi-layer dilated convolution is removed, a moderate reduction in AP is observed, indicating the significance of the model's receptive field for prediction capabilities. 
  \item \textbf{Model C} Replacing the dynamic GCN with a linear layer results in a substantial drop in overall performance, as the model loses its ability to capture spatial relationships among targets, relying solely on video features for inference. 
  \item \textbf{Model D} Substituting the adaptive adjacency matrix with a fully connected matrix still allowes the graph neural network to be constructed, but led to a reduction in the model's spatial learning capability, resulting in a decrease in AP.
\end{itemize}

While GRU \cite{chung2014empirical} has demonstrated its effectiveness in temporal sequence learning, other models such as LSTM \cite{van2020review}, Transformer \cite{vaswani2017attention} and TCN \cite{bai2018empirical} also exhibit comparable capabilities in long-term sequence modeling. Therefore, we conduct experiments to explore the performance differences of various temporal modeling modules in accident anticipation. The experimental results are presented in Table~\ref{GRU}.

The experimental results indicate that each model exhibits distinct strengths within our accident anticipation framework. LSTM achieves the highest accuracy, whereas GRU attains the longest mTTA. Transformer demonstrates the most balanced overall performance, while TCN consumes the least computational resources without a noticeable performance drop. 

Thus, under the current framework, multiple long-term sequence modeling architectures can be selected based on specific requirements. For applications prioritizing high accuracy or extended accident alert time, LSTM or GRU are preferable. If computational resources are constrained-a common scenario in vehicle-mounted platforms-TCN serves as a more efficient choice. For a balanced trade-off across multiple performance metrics, Transformer stands out as the optimal option.

\subsection*{Visualization and Qualitative Analysis}

To better measure the performance of our proposed model and compare it with previous state-of-the-art models, we visualize some scenes in the dataset and compare the anticipations of our model with another state-of-the-art model.

First, we show a detailed scenario analysis, as shown in Figure \ref{Fig_visual}a. In this scenario, the accident participants only appear in the middle of the video, and then the accident occurs. We provide detailed annotations of the time before the accident is predicted, the time from the accident anticipation to the accident occurrence, and the performance comparison between models. Our accident confidence trend is consistent with the key information in the scenario, when the accident participants come into view, the system successfully predicts the impending accident.

Figure \ref{Fig_visual}b shows a positive case from DAD testing set. This video is very blurry due to heavy rain and reflections from the windshield. In this case, DSTA failed to use visual features to make a successful prediction due to environmental interference, while our model successfully predicted the impending accident by modeling the spatial position between objects. At the end of the video, the motorcycle involved in the accident is blocked by the car after the accident, so the position and visual information are lost, resulting in a rapid drop in the subsequent accident confidence.

Figure \ref{Fig_visual}c shows an arduous negative case from DAD testing set. In this case, the bus and the car were very close to each other. If we only consider the pixel distance between the two, it is easy to misjudge that an accident is about to happen. However, our model uses the relative speed, relative distance, and visual features between the two vehicles for reasoning. Therefore, when DSTA misjudges, our model does not produce false predictions.

 Figure \ref{Fig_visual}d shows another negative case. In this video, the ego car is waiting for a traffic light. Therefore, the objects that occupy the main part of the field of view at a close distance are stationary, while the objects at a long distance are mostly blurred or overlapped. Besides, there are a ton of closely parked motorcycles in this video. This case once again demonstrates the advantage of our method over traditional methods, that is, it considers multiple factors and excludes motorcycles without drivers.instead of relying on a single feature.

 Figure \ref{Fig_visual}e is a typical example of a failure in our accident anticipation system. In this case, both our method and DSTA produced incorrect accident anticipations. The possible reason is that the car and motorcycle parked on the side are too close, and as the ego vehicle moves, the camera's perspective changes to a certain extent, causing the two to appear to be getting closer and eventually overlapping. This exposes a problem with current accident prediction methods, which is that it is difficult to handle the impact of perspective changes caused by the movement of the ego vehicle on the prediction results. Although we have tried our best to consider more relevant factors, there are still some cases where anticipations fail.

\section*{Conclusion}
In this study, we propose a traffic accident anticipation framework based on dynamic graph neural network, and release a dataset named Anticipation of Traffic Accident (AoTA), which contains a rich set of traffic accident samples with comprehensive annotations. Furthermore, we introduce a world model-driven scene generation framework for camera video data augmentation, successfully demonstrating its effectiveness in this task. Our proposed model exhibited outstanding performance across key metrics on multiple datasets, establishing new benchmarks in this field. Our research has verified that generated data has a positive effect in specific tasks. In future research, we should consider generating reasonable accident scenarios by analyzing accident principles and negative scenarios. This not only relies on the understanding of accident factors and the relationship between text and video, but also requires building a traffic scenario generation and simulation model that can generate real accident scenarios. In addition, the generated videos are somewhat different from real-world videos in terms of target coherence, noise, etc. By bridging this gap, we can connect traffic scene generation with existing autonomous driving systems, effectively alleviating the current problems of insufficient data and annotations.

\section*{Acknowledgment}
This work was supported by the Science and Technology Development Fund of Macau [0122/2024/RIB2 and 001/2024/SKL], the Research Services and Knowledge Transfer Office, University of Macau [SRG2023-00037-IOTSC, MYRG-GRG2024-00284-IOTSC], the Shenzhen-Hong Kong-Macau Science and Technology Program Category C [SGDX20230821095159012], the State Key Lab of Intelligent Transportation System [2024-B001], and the Jiangsu Provincial Science and Technology Program [BZ2024055].

\section*{DATA availability}
All data and the proposed dataset of this study are available on GitHub (https://github.com/humanlabmembers\\/Anticipation-of-Traffic-Accident) .

\section*{CODE availability}
The code used to generate the results of this study is available on GitHub (https://github.com/humanlabmembers\\/Anticipation-of-Traffic-Accident) .

\section*{Author Information }
Yanchen Guan: Conceptualization of this study, Method
ology, Experiment, Writing. Haicheng Liao: Conceptual
ization of this study, Writing. Chengyue Wang: Writing.
Xingcheng Liu: Experiment. Jiaxun Zhang: Experiment.
Zhenning Li: Conceptualization of this study,Writing.

\section*{Ethics declarations}
The authors declare no competing interests.

\bibliographystyle{unsrt}
\bibliography{cas-refs}

\newpage

\begin{table*}[t]
    \centering
    \caption{Comparison between AoTA dataset and existing datasets. Information about DAD, CCD and A3D is obtained from their
released sources. \textit{Accident Time} means the starting time of the accident is fixed at a specific frame or randomly distributed. \textit{Ego-Involved} represents the ego-vehicles are involved in some accidents. \textit{Day/Night} indicates the accident scenes occur both during the day and at night. }
    \setlength\tabcolsep{2pt}
     \resizebox{0.8\linewidth}{!}{
    \begin{tabular}{cccccccccc}
    \hline
        Dataset & Videos & Positives & Frames& FPS &Accident Time  &Ego-Involved &Accident Reasons&Weather&Day/Night\\ \hline
        DAD \cite{chan2017anticipating}& 1750 & 620  &100& 20 & Fixed &\checkmark &~ &~ &~\\ 
        CCD \cite{BaoMM2020}& 4500 & 1500  &50& 10 & Random &\checkmark &~ &\checkmark &\checkmark\\ 
        A3D \cite{yao2019unsupervised}& 1500 & 1500  &100& 20 & Random  &\checkmark &\checkmark &~ &~\\ 
        AoTA (Ours) & 4800 & 3600  &50& 10 & Random &\checkmark &\checkmark &\checkmark &\checkmark\\
    \hline
    \end{tabular}
    }
    
    \label{datasets}
\end{table*}

\begin{table*}[h]
\centering
\caption{\small{Comparison of models seeking \textbf{balance} between mTTA and AP on DAD, A3D, AoTA and AoTA+. \textbf{Bold} and \underline{underlined} values represent the best and second-best performance. Instances where values are not available are marked with a dash (``-'').}}
\setlength\tabcolsep{2pt}
 \resizebox{0.8\linewidth}{!}{
\begin{tabular}{lcccccccccc}
\hline \specialrule{0em}{1pt}{1pt}
\multirow{2}[2]{*}{Model} & \multicolumn{2}{c}{DAD}& \multicolumn{2}{c}{A3D} & \multicolumn{2}{c}{AoTA} & \multicolumn{2}{c}{AoTA+} & \multicolumn{2}{c}{DAD (Best AP)} \\
\cmidrule(lr){2-3} \cmidrule(lr){4-5} \cmidrule(lr){6-7} \cmidrule(lr){8-9}\cmidrule(lr){10-11}
 & AP (\%) & mTTA (s)& AP (\%) & mTTA (s) & AP (\%) & mTTA (s) & AP (\%) & mTTA (s) & AP (\%) & mTTA (s) \\
\hline
DSA \cite{chan2017anticipating}& 48.1 & 1.34  & 92.3 & 2.95 & - & - & - & -&63.5&1.67 \\
AdaLEA \cite{suzuki2018anticipating}& 52.3 & 3.43  & 92.9 & 3.16 & - & - & - & -&52.3&3.43 \\
UString \cite{BaoMM2020}& 53.7 & 3.53  & 93.2 & \textbf{3.24} & 71.2 & 2.30 & 72.4 & 1.71&68.4&1.63\\
DSTA \cite{karim2022dynamic}& 56.1 & \underline{3.66}  & 93.5 & 2.87 & \underline{72.6} & 1.95 & \underline{73.6} & 1.98&72.3&1.52 \\
GSC \cite{wang2023gsc}& 60.4 & 2.55  & 94.9 & 2.62 & - & - & - & -&68.9&1.33 \\

AccNet \cite{liao2024real} & 60.8 & 3.58  & \underline{94.9} & 2.62 & 71.8 & \underline{2.31} & 73.5 & \underline{2.37} &70.1&1.73\\

THAT-NET \cite{liu2023net}& 77.8 & 3.64 & - & - & - & - 
 & - & - &77.8&\textbf{3.64}\\ 
MASTTA \cite{patera2025multi}& \underline{80.8} & 3.32 & - & - & - & - 
 & - & - &\underline{80.8}&3.32\\ \hline

\textbf{Ours} & \textbf{83.2} & \textbf{3.99}  & \textbf{95.1} & \underline{3.18} & \textbf{75.4} & \textbf{3.46} & \textbf{78.7} & \textbf{3.40} &\textbf{86.3}&\underline{3.58}\\
\hline
\end{tabular}
}
\label{table:balance}
\end{table*}

\begin{table}[htbp]
\centering
\caption{\small{Comparison of our model seeking \textbf{best AP} on DAD datasets with different level data augmentation. All experimental results are tested on \textbf{original DAD testing set}.}}
 \resizebox{0.9\linewidth}{!}{
\begin{tabular}{lccc}
\hline 
Training Dataset & Number of Videos & AP (\%)  & mTTA (s) \\
\hline 
DAD & 1280 & 86.3 & 3.58 \\
DAD (+10\%) & 1360 & 88.7$_{\uparrow{2.8\%}}$ & 4.05$_{\uparrow{13.1\%}}$ \\
DAD (+20\%) & 1440 & 89.2$_{\uparrow{3.4\%}}$ & 4.03$_{\uparrow{12.5\%}}$ \\
DAD (+30\%) & 1520 & 89.5$_{\uparrow{3.7\%}}$ & 3.96$_{\uparrow{10.6\%}}$ \\
DAD (+40\%) & 1600 & 92.1$_{\uparrow{6.7\%}}$ & 3.71$_{\uparrow{3.6\%}}$ \\
DAD (replaced 40\%) & 1280 & 84.2$_{\downarrow{2.4\%}}$ & 3.30$_{\downarrow{7.8\%}}$ \\
\hline
\end{tabular}
}

\label{table:dad+}
\end{table}

\begin{table}[htbp]
\centering
\caption{Ablation studies of different modules on DAD dataset.
DC, DGCN, ADJ represent multi-layer dilated convolution, dynamic GCN, and adaptive adjacency matrix, respectively.}
 \resizebox{0.9\linewidth}{!}{
\begin{tabular}{ccccccc}
\hline \specialrule{0em}{1pt}{1pt}
MODEL & GRU & DC & DGCN & ADJ & AP (\%) & mTTA (s) \\ \hline
A &  & \checkmark & \checkmark & \checkmark & 50.2$_{\downarrow{39.7\%}}$ & 3.62$_{\downarrow{9.3\%}}$ \\ 
B & \checkmark &  & \checkmark & \checkmark & 57.4$_{\downarrow{31.0\%}}$ & 4.04 \\ 
C & \checkmark & \checkmark &  & \checkmark & 48.5$_{\downarrow{41.7\%}}$ & 3.54$_{\downarrow{11.3\%}}$ \\ 
D & \checkmark & \checkmark & \checkmark &  & 77.4$_{\downarrow{7.1\%}}$ & 4.07 \\ \hline
Final & \checkmark & \checkmark & \checkmark & \checkmark & 83.2 & 3.99 \\ \hline
\end{tabular}
}

\label{Ablation}
\end{table}

\begin{table}[htbp]
\centering
\caption{Comparison between different models in accident anticipation time series modeling. \textbf{Bold} represents the best results of different models in this item. The input and output dimensions and hidden layer dimensions of each model are consistent.}
 \resizebox{1.0\linewidth}{!}{
\begin{tabular}{ccccccc}
\hline \specialrule{0em}{1pt}{1pt}
\multirow{2}[2]{*}{Model} & \multicolumn{2}{c}{Balance}& \multicolumn{2}{c}{Best AP} & \multirow{2}[2]{*}{FLOPs (G)} &\multirow{2}[2]{*}{Params (M)} \\
\cmidrule(lr){2-3} \cmidrule(lr){4-5} 
 & AP (\%) & mTTA (s)& AP (\%) & mTTA (s) \\
\hline
GRU & 83.2 & \textbf{3.99} & 86.3 & 3.58 & 2.40 & 2.40 \\ 
LSTM & 84.6 & 3.75 & \textbf{90.2} & 3.31 & 3.19 & 3.18 \\ 
Transformer & \textbf{85.1} & 3.95 & 89.5 & \textbf{3.84} & 2.19 & 2.18 \\ 
TCN & 84.4 & 3.49 & 87.5 & 3.36 & \textbf{1.61} & \textbf{1.61} \\ \hline
\end{tabular}
}
\label{GRU}
\end{table}

\end{document}